\documentclass{cls/ieee-ojvt}

\usepackage{cite}
\usepackage{amsmath,amssymb,amsfonts}
\usepackage{algorithmic}
\usepackage{algorithm}
\usepackage{graphicx,color}
\usepackage{textcomp}
\usepackage{stfloats}
\usepackage{colortbl}
\usepackage{multirow} 
\usepackage{stmaryrd}
\usepackage[numbers]{natbib}
\usepackage[acronym]{glossaries}
\usepackage{hyperref}
\hypersetup{
    colorlinks=true,
    linkcolor=blue,
    citecolor=blue,
    urlcolor=magenta
}

\def\BibTeX{{\rm B\kern-.05em{\sc i\kern-.025em b}\kern-.08em
    T\kern-.1667em\lower.7ex\hbox{E}\kern-.125emX}}
\AtBeginDocument{\definecolor{ojcolor}{cmyk}{0.93,0.59,0.15,0.02}}
\def\OJlogo{\vspace{-12pt}\includegraphics[height=24pt]{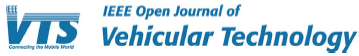}}
\makeglossaries

\begin{document}
\receiveddate{XX Month, XXXX}
\reviseddate{XX Month, XXXX}
\accepteddate{XX Month, XXXX}
\publisheddate{XX Month, XXXX}
\currentdate{17 June, 2025}
\doiinfo{OJVT.2024.0627000}

\title{FIN: Fast Inference Network \\ for Map Segmentation}
\author{RUAN~BISPO\textsuperscript{1,2,3}~(Member,~IEEE),
        TIM~BROPHY\textsuperscript{1,2,3}~(Member,~IEEE),
        REENU~MOHANDAS\textsuperscript{1,2,3},
        ANTHONY~SCANLAN\textsuperscript{1,2,3},
        CIARÁN~EISING\textsuperscript{1,2,3}~(Senior Member,~IEEE)
}
        
\affil{Department of Electronic and Computer Engineering, University of Limerick, Limerick V94 T9PX, Ireland}
\affil{Data Driven Computer Engineering Research Centre, University of Limerick, Limerick V94 T9PX, Ireland}
\affil{Lero, The Irish Software Research Centre, University of Limerick, Limerick V94 NYD3, Ireland}
 
\corresp{CORRESPONDING AUTHOR: Ruan~Bispo (e-mail: bispodos.ruan@ul.ie).}
\authornote{This work was supported by Lero, (the Research Ireland Centre for Software) under Grant 13/RC/2094\_P2.}
\markboth{FIN: Fast Inference Network for Map Segmentation}{Bispo et al.}

% Acronyms List

\newacronym{fin}{FIN}{Fast Inference Network for Map Segmentation}
\newacronym{pan}{PAN}{Pillars-attention-based Network}
\newacronym{bev}{BEV}{Bird's-Eye View}
\newacronym{miou}{mIoU}{mean Intersection over Union}
\newacronym{iou}{IoU}{Intersection over Union}
\newacronym{fps}{FPS}{Frames per second}
\newacronym{mfa}{MFA}{Multi-modal Feature Aggregation}
\newacronym{mdca}{MDCA}{Multi-modal Deformable Cross-attention}
\newacronym{rvt}{RVT}{Radar-assisted View Transformation}

\begin{abstract}
Multi-sensor fusion in autonomous vehicles is becoming more common to offer a more robust alternative for several perception tasks. This need arises from the unique contribution of each sensor in collecting data: camera-radar fusion offers a cost-effective solution by combining rich semantic information from cameras with accurate distance measurements from radar, without incurring excessive financial costs or overwhelming data processing requirements. Map segmentation is a critical task for enabling effective vehicle behaviour in its environment, yet it continues to face significant challenges in achieving high accuracy and meeting real-time performance requirements. Therefore, this work presents a novel and efficient map segmentation architecture, using cameras and radars, in the \acrfull{bev} space. Our model introduces a real-time map segmentation architecture considering aspects such as high accuracy, per-class balancing, and inference time. To accomplish this, we use an advanced loss set together with a new lightweight head to improve the perception results. Our results show that, with these modifications, our approach achieves results comparable to large models, reaching 53.5 mIoU, while also setting a new benchmark for inference time, improving it by 260\% over the strongest baseline models.

\end{abstract}

\begin{IEEEkeywords}
BEV, camera-radar, \textit{nuScenes}, map segmentation, sensor fusion, real-time.
\end{IEEEkeywords}

\maketitle
\section{INTRODUCTION}\label{sec:introduction}

With the increasing imperative for advances in the automated vehicles domain \cite{zwetsloot2020vision}, various perception tasks have seen constant improvements, including map segmentation, which plays a key role in distinguishing navigable from non-navigable areas, as misclassifications can lead to unsafe manoeuvres or collisions with obstacles and vulnerable road users. Unlike traditional image segmentation, map segmentation projects sensor data onto a map grid, commonly with the shape of 100~m~\(\times\)~100~m, with the precision of 0.5~m for each cell. For this task, each cell receives a semantic label such as drivable surface, pedestrian crossing, sidewalk, stop line, carpark, road divider, or lane divider \cite{schramm2024bevcar, zou2024unim}. This unified representation simplifies downstream modules for some tasks, such as path planning and trajectory prediction, by providing a holistic, metric-grounded view of the environment \cite{liu2022bevfusion}.

Based on prior work, camera-only methods face significant challenges when reconstructing fine-grained details: objects with thin or elongated shapes (e.g., dividers, sidewalks) may occupy only a few pixels in the image plane or be occluded by other scene elements, resulting in fragmented or missing segments in the map. To mitigate these issues, some approaches have fused LiDAR, whose accurate depth measurements improve spatial completeness and structural recovery, but at the cost of expensive sensors and sensitivity to adverse weather conditions \cite{liu2022bevfusion}. Moreover, camera–radar fusion has emerged as an attractive alternative, leveraging radar’s robustness to weather variation and its direct depth and velocity measurements to complement visual information and better recover small or partially occluded objects in the \acrfull{bev} representation \cite{schramm2024bevcar}.

Another key challenge related to the map segmentation task is achieving a real-time application without sacrificing precision. This characteristic is essential to guarantee the safety of both vehicle occupants and vulnerable road users outside the vehicle in real-world applications \cite{schramm2024bevcar}. Mapping delays or spatial inaccuracies can lead to unsafe trajectory planning and prediction, resulting in mismatches between drivable surfaces and environments with pedestrians, cyclists, vehicles, or other critical objects. Thus, modern architectures must have a high semantic accuracy with low-latency results, typically under 50 ms per frame, to support safety-critical decision making in dynamic traffic scenarios \cite{sze2024real}.  

To overcome these limitations, we propose a novel camera-radar fusion architecture, \acrfull{fin}, that integrates both sensor inputs for robust and real-time map semantic segmentation in the \acrshort{bev} domain. Our design builds on the principles of multi-scale simplified networks: a ResNet-50 backbone extracts features from images, while radar features are encoded into complementary \acrshort{bev} feature maps using \acrshort{pan} \cite{bispo2025panpillarsattentionbasednetwork3d}. Next, we fuse these features using cross-attention and a simplified U-Net-based head that refines the feature maps using a multi-scale approach. Finally, we apply a set of losses to improve our metrics and per-class results, yielding a unified representation that preserves fine-grained details without incurring excessive computational costs.

In this study, we conduct experiments on the nuScenes validation set, as well as on a filtered version of this dataset using specific weather and lighting conditions. Our results demonstrate that FIN overcomes the baselines in map segmentation accuracy, measured by mean Intersection-over-Union (mIoU), while running at approximately 38.5 ms or 26 \acrfull{fps} on a single NVIDIA A100 GPU, representing a 260\% speed-up over our strongest baseline \cite{schramm2024bevcar}. We further ablate the contributions of our proposed head and loss functions, designed to emphasise boundary and balanced per-class results, besides showing their impact on our results.
% \subsection{SUMMARY OF OUR MAIN CONTRIBUTIONS}

The main contributions of this paper are as follows: 
\begin{enumerate}
    \item We propose \acrshort{fin}, a real-time camera–radar sensor fusion architecture for map segmentation, achieving approximately 26 \acrshort{fps} with ResNet-50 as the camera backbone.
    \item We are proposing a new map segmentation head using a simplified U-Net to enhance the feature map through a \acrshort{bev} multi-scale feature extractor.
    \item We propose a new set of losses to improve results, including losses adapted from the medical domain, proposing a normalised version of one of them to improve the results for the map segmentation task, while the baseline uses only a single loss for this task.
    \item We benchmark our method against heavier image backbones (ViT-B/14, ResNet-101), showing that \acrshort{fin} with ResNet-50 not only surpasses them in accuracy but also runs 260\% faster than the baseline.
\end{enumerate}

The rest of this paper is organised as follows: Section \ref{sec:related_work} provides an overview of related work in the fields of multi-sensor fusion and \acrshort{bev} map segmentation. Section \ref{sec:architecture} describes the full proposed \acrshort{fin} architecture, including its sensor inputs and feature extractor, network design, the novel lightweight head, and the loss set proposal. Section \ref{sec:methodology} details the experimental setup and evaluation metrics. Section \ref{sec:results} presents our quantitative and qualitative results. Finally, Section \ref{sec:conclusion} concludes the paper and discusses potential paths for future research.

\section{RELATED WORK}\label{sec:related_work}
Recent \acrshort{bev}-based perception architectures, including the map segmentation task on nuScenes, can be largely grouped into two categories, according to the sensing used: camera‑only, and sensor fusion methods \cite{li2022bevformer}. Camera‑only approaches such as CVT \cite{zhou2022cross}, OFT \cite{roddick2018orthographic}, LSS \cite{philion2020lift}, M²BEV \cite{xie2022m}, BEVFormer \cite{li2022bevformer} and UniM²AE \cite{zou2024unim} exploit multi‑view images exclusively, achieving competitive \acrshort{miou} but often suffering from problems in the camera to \acrshort{bev} transformation, especially for objects further away from the vehicle, or under adverse weather conditions \cite{schramm2024bevcar}. In contrast, sensor fusion methods, more specifically camera-radar methods, exemplified by BEVCar \cite{schramm2024bevcar}, Simple‑BEV++ \cite{harley2023simple} and BEVGuide \cite{man2023bev}, have gained traction for their cost‑benefit balance, improving robustness in low‑visibility while maintaining relatively modest hardware requirements. 
% From this, it is possible to observe that fusion methods have better metrics and robustness against low visibility, far, and occluded o

Inference time is also a critical constraint for real‑time autonomous driving. Methods built on heavyweight backbones like ViT‑B/14 report only around 3.6 \acrshort{fps} on nuScenes \cite{harley2023simple}, while ResNet‑101 variants such as BEVCar operate at approximately 7.3 \acrshort{fps} \cite{schramm2024bevcar}. BEVFusion with ResNet‑50 attains good results, around 8.4 \acrshort{fps} \cite{liu2022bevfusion}, but still below real‑time thresholds. These approaches presented state-of-the-art metrics in terms of \acrshort{miou} when they were published, but they had limitations related to this inference time restriction. Our proposed \acrshort{fin}, by contrast, achieves approximately 26 \acrshort{fps} using ResNet‑50 as the image backbone, meeting stringent deployment requirements for real‑world applications \cite{sze2024real}.

Additionally, most prior works employ a relatively simple decoder for the map segmentation head or even lightweight transformer-based architectures to predict the \acrshort{bev} segmented maps, which often lack a balanced multi‑scale feature aggregation. For instance, BEVCar’s convolutional‑based head focuses on high‑level semantics but misses fine spatial details \cite{schramm2024bevcar}, and the respective BEVFusion module applied is also based only on a one-scale convolutional module, using a vanilla convolutional network \cite{liu2022bevfusion}. Additionally, PON \cite{roddick2020predictingsemanticmaprepresentations}, and UniM²AE \cite{zou2024unim} introduce transformer-based architectures to address this task but incur significant overhead. 
% These approaches can limit performance on classes requiring both context and detail.

Many methods for the map segmentation task rely exclusively on cross‑entropy loss, or a variation, such as focal loss (e.g., BEVCar \cite{schramm2024bevcar}, BEVFormer \cite{li2022bevformer}, Simple‑BEV++ \cite{harley2023simple}, BEVFusion \cite{liu2022bevfusion}). However, recent works \cite{hurtado2024semanticscenesegmentationrobotics, 10588555} have highlighted the advantages of hybrid losses, for example, combining Dice loss, which is well suited for handling class imbalance applied to natural language processing tasks \cite{li2019dice}, or focal‑based losses \cite{li2023fb}. We extend this approach by proposing an advanced loss set, which also includes a normalised variant of the Boundary loss \cite{kervadec2018boundary}. This loss was first proposed to recover small regions in segmented ultrasound images. We are proposing its adaptation to the \acrshort{bev} domain as we have observed the same problem of coarse boundaries in the segmentation. This advanced loss set further improves overall metrics and enhances boundary segmentation by yielding better performance per class.

Moreover, real-world deployment demands robustness to weather, lighting, and sensor noise. Prior studies highlight the strengths of sensor fusion methods under low-light \cite{liu2022bevfusion, schramm2024bevcar}, showing better resilience in different weather conditions. However, only a few works systematically assess safety margins or metric degradation in real-world applications. Our \acrshort{fin} architecture includes additional safety analyses across varied nuScenes conditions and distances, demonstrating variations in \acrshort{miou} and highlighting our limitations under these scenarios.

\section{ARCHITECTURE}\label{sec:architecture}
\begin{figure*}[!t]
\centering
\includegraphics[width=7in]{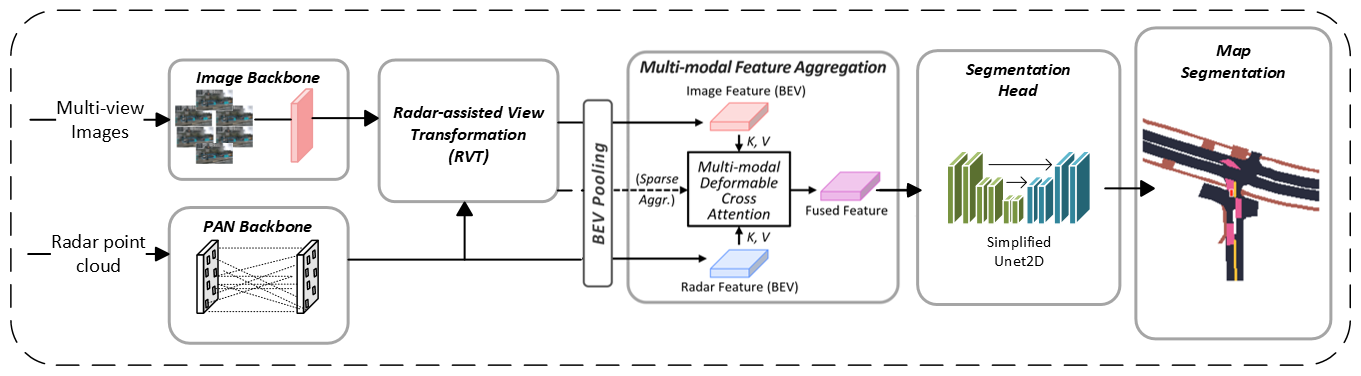}
\caption{The overall architecture of \acrshort{fin} is as follows. First, radar and camera features are extracted in parallel using their respective backbones. Second, using the \acrshort{rvt} module, we convert these features into a \acrshort{bev} representation. Third, the Multi-Modal Feature Aggregation module merges and refines the radar and camera \acrshort{bev} features, which are then fed into the Segmentation Head.}
\label{fig:FIN_overview}
\end{figure*}

In this section, we present our proposed \acrshort{fin} architecture for map segmentation, where we are using a camera-radar fusion network to predict the map segmentation task. Our \acrshort{fin} architecture uses lightweight blocks using a computationally efficient approach to extract features from different sources and refine them into the \begin{math}200\times200\end{math} segmentation grid map, with 0.5 m of resolution, by merging attention and multi-scale convolutional networks. Additionally, we are using a set of losses commonly used in the medical and 3D occupancy prediction domains to improve our results and address issues related to the shape of objects and accuracy.

As shown in Figure \ref{fig:FIN_overview}, our input data are the multi-view camera images and the radar point cloud. Feature fusion methods can generally be divided into early, middle, and late fusion, depending on the stage at which data from different modalities are combined \cite{shi2024radarcamerafusionobject}. In our architecture, we are using a middle fusion approach as it balances the preservation of modality-specific features with the benefits of joint representation learning \cite{wang2023multi}. We first extract the features for the camera and radar branches in parallel using two different backbones. In sequence, we apply the \acrfull{rvt} to convert the image features from a camera perspective into a \acrshort{bev}, followed by an attention-based fusion method to merge and refine the \acrshort{bev} features. Finally, we are using a novel U-Net-based segmentation head to transform from the \acrshort{bev} features to the map predictions.

In the following subsections, each component of the diagram in Fig. \ref{fig:FIN_overview} is described, considering each sub-block used in our architecture. We first describe the image and radar backbones, our first stage of feature extraction. Secondly, we present the components responsible for the view transformation, converting our features to a unified \acrshort{bev} space. Thirdly, we show the fusion method, considering the multi-modal \acrshort{bev} features, and its feature enhancement block. Finally, we detail our proposed simplified segmentation head and the set of losses used in our architecture.

\subsection{IMAGE BACKBONE}

For the multi-image feature extractor, we use ResNet, aggregating all six images per frame to feed our image backbone. We adopt a method entirely based on convolutional neural networks as the image backbone, due to their main advantage of delivering good results with lower computational cost compared to approaches based on vision transformers, which typically require significantly more computational resources \cite{jain2024comparative}. In this work, we choose to use ResNet-50 as it represents a balanced option, although it offers slightly lower accuracy compared to more commonly used variants in the literature, such as ResNet-101; it provides a lightweight architecture that facilitates architectural refinements in subsequent phases, without compromising inference time.

\subsection{PAN BACKBONE}

The \acrshort{pan} backbone \cite{bispo2025panpillarsattentionbasednetwork3d} is employed as the feature extraction module for radar data. In this setup, the input consists of a multi-sensor point cloud obtained from five radar sensors positioned to cover a full 360-degree field of view. The output of this module is a set of extracted features that represent the spatial and contextual information within the radar data. Initially, the raw point cloud is transformed into pillar-based features using a pillar network (PillarNet) \cite{lang2019pointpillars}, which transforms the data into a structured grid format. The pillar representation is composed of voxelized features arranged in an \begin{math}H \times W\end{math} grid, where the vertical (third) spatial dimension is collapsed. This dimensionality reduction is performed to reduce computational overhead while enhancing the efficiency of feature extraction for 2D radar data (considering the spatial dimension), which has limited angular resolution to map the 3D environment.

The pillar-based features are first passed through an MLP-based encoder module, which serves to reduce feature sparsity and transform the raw input into a more compact representation suitable for subsequent processing. These encoded features are then fed into an attention module designed to enhance the features by focusing exclusively on non-empty pillars, thereby improving the relevance of the information being propagated by these pillars. Following the attention mechanism, the refined features are transformed back to pillars and forwarded to the convolutional layers of the PAN backbone, a two-stage convolutional block, which enhances local feature learning while also adjusting the spatial dimensions of the output to prepare them for the next components.

\subsection{RADAR-ASSISTED VIEW TRANSFORMATION}

After feature extraction from both the camera and radar backbones, the next step in our architecture involves the camera-view to \acrshort{bev} transformation. We adopt the \acrshort{rvt} proposed by \citeauthor{kim2023crn} \cite{kim2023crn}, which combines radar and camera data to enhance the accuracy of the transformation. This method projects features from the camera perspective into the \acrshort{bev} space by leveraging radar-based depth estimation. Specifically, it incorporates both radar occupancy and depth distribution to generate a more accurate depth map. 

Additionally, fusing the precise depth information from radar with the semantic features of the camera images, the model improves the reliability of the camera-based depth estimation. Following this, the \acrshort{bev} Pooling transformation is applied to the camera and radar features using a CUDA-accelerated Voxel Pooling operation \cite{li2023bevstereo}, which unifies multi-modal features within the \acrshort{bev} representation. In our case, the \acrshort{pan} backbone \cite{bispo2025panpillarsattentionbasednetwork3d} provides more refined feature maps and improved occupancy estimation, allowing us to inject more precise spatial information into the transformation stage, enhancing the overall quality and performance of the \acrshort{bev} representation.

\subsection{MULTI-MODAL FEATURE AGGREGATION}

Once the features have been projected into the \acrshort{bev} space and unified through voxel pooling, the next step in our architecture is the fusion of these multi-modal representations. This is performed during the \acrfull{mfa} stage, which aims to effectively integrate the complementary strengths of radar and camera features, in addition to reducing calibration and data misalignment errors. To achieve this, we employ the \acrfull{mdca} mechanism proposed by Kim \textit{et al.} \cite{kim2023crn}. 

The \acrshort{mdca} module utilises a cross-attention strategy that dynamically aligns and merges information from both modalities, focusing particularly on spatial regions where their features are most correlated. By doing so, it mitigates common issues related to sensor calibration errors and spatial misalignment between radar and camera inputs. The \acrshort{mdca} operation is formally defined by \eqref{eq:cross_att}:

\begin{multline}
\label{eq:cross_att}
\mathrm{MDCA} = \\
\sum_{h=1}^{H} W_{h} \left[
    \sum_{m=1}^{M} \sum_{k=1}^{K}
    A_{hmqk} \cdot W'_{hm} \,
    x_{m} \big( \phi_{m}(p_{q} + \Delta p_{hmqk}) \big)
\right]
\end{multline}

In this formulation, $H$ denotes the number of attention heads, $M$ the number of modalities, and $K$ the number of sampling points. $W_{h}$ and $W'_{hm}$ are the output and input projection matrices, respectively, for head $h$ and modality $m$. The attention weights $A_{hmqk}$ indicate the importance of each sampled feature, while $\phi_{m}$ scales the normalised coordinates $p_{q}$ for each query element in case two modalities have different shapes. The term $\Delta p_{hmqk}$ represents learned offsets that enable flexible sampling around the reference points, improving cross-modal alignment. This deformable structure enables flexible, content-aware fusion that effectively reduces alignment inconsistencies between modalities. After the cross-attention fusion, the aggregated features are passed through a convolutional layer, following the ResNet-18 base model. These layers serve to refine the multi-modal features by enhancing spatial coherence and semantic consistency.

\subsection{MAP SEGMENTATION HEAD}

\begin{figure}[!htbp]
    \centering
    \includegraphics[width=3.4in]{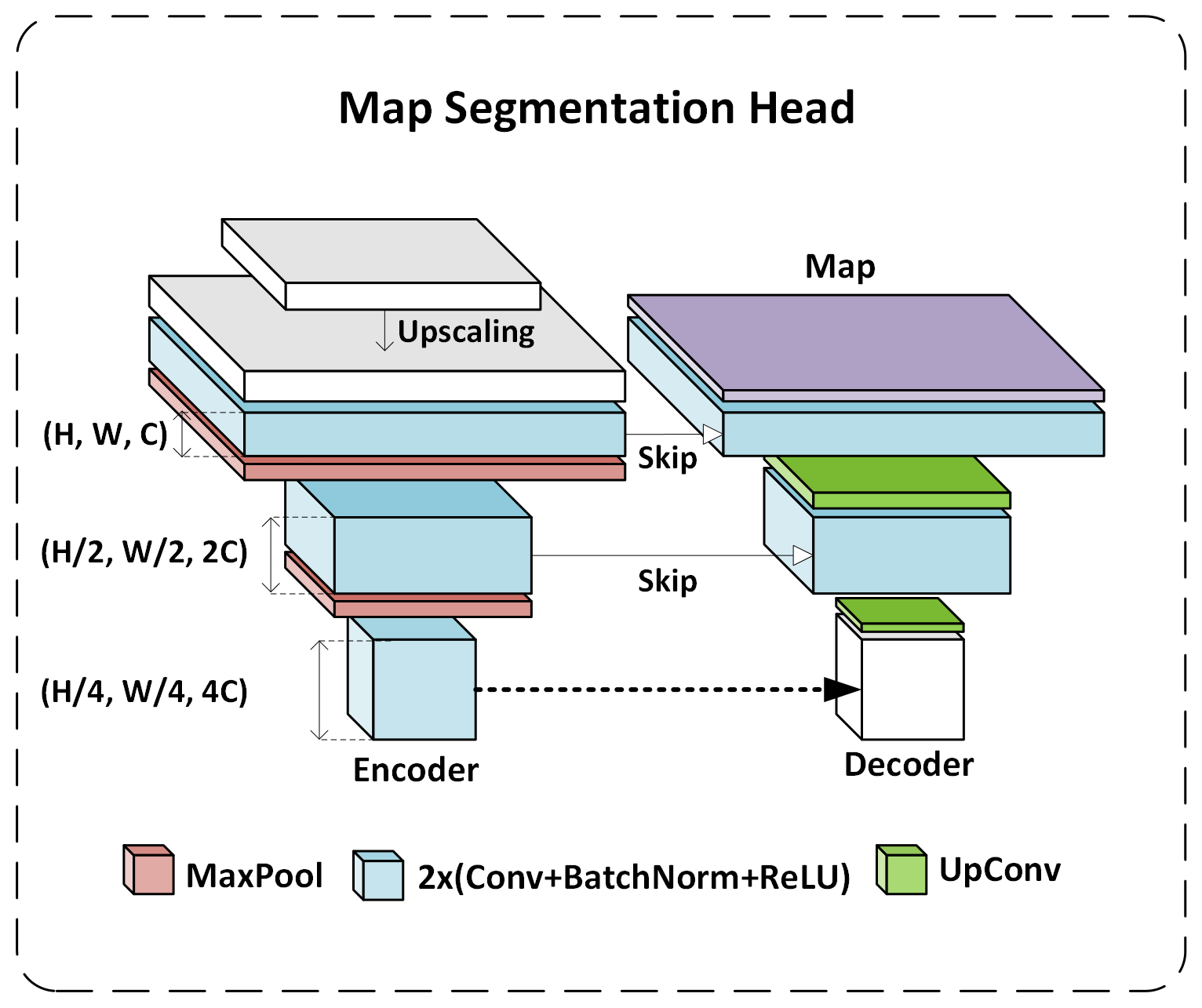}
    \caption{Map Segmentation Head showing how the \acrshort{bev} features are transformed into the map segmentation, where H, W, and C are the \acrshort{bev} height, width, and channels, respectively. The skip connections are concatenated to the UpConv features before the convolutional block, except at the bottleneck stage, where no upsampled features exist; in that case, the features are simply forwarded (white box) without concatenation.}
    \label{fig:map_seg_head}
\end{figure}

For the map segmentation head, we are applying a 2-stage convolutional network, inspired by an architecture proposed in \cite{ronneberger2015u}. This method utilises different scales of features to capture spatial information, effectively locating the shape of the object within the scene, which improves accuracy and adaptability to varying object sizes. In our head, firstly, we are applying the U-Net concept for the \acrshort{bev} space, different from the literature \cite{schramm2024bevcar} that uses a vanilla convolutional layer to enhance performance. Secondly, we are also adapting the U-Net architecture, reducing the number of stages and using a last convolutional layer to adjust the map segmentation output. A summary of this application is shown in Figure \ref{fig:map_seg_head}.

In the encoder branch, the \acrshort{bev} features are passed through a bilinear up-scaling method to convert the \acrshort{bev} grid from \(128\times128\), the standard feature shape, to the expected map segmentation shape, \(200\times200\). After this stage, we employ a two‐level U-Net module, where each level comprises a pair of successive convolutional layers, followed by batch normalisation and an activation function, composing a lightweight variant of the classical four-stage U-Net (which typically reaches up to 1024 feature channels). By reducing the number of intermediate feature maps and limiting the network depth to two encoder–decoder scales, we decrease computational complexity and memory footprint to fit in our real-time application. At the end of the encoder, the features tensor is downsampled to a spatial resolution of \(50\times50 \), and the final channel dimensionality is 256, thereby achieving a balance between spatial detail and high-level semantic abstraction.

After the encoder branch, we apply the decoder to transform the features tensor back to the original dimension of \(200\times200\). Firstly, we use the last extracted features, dimension \(H/4 \times W/4 \times 4C\), to feed an UpConv block, where we are using a 2D transposed convolution operator with the same number of channels as the input. In sequence, we use the upsampled feature maps concatenated with the equivalent features from the encoder branch, as shown in Fig. \ref{fig:map_seg_head} by the skip arrows, to feed a convolutional block, using the same structure used in the encoder branch. This operation is repeated until the size of the feature maps becomes \(200\times200\). The last convolutional layer outputs a number of channels equal to the desired number of classes.

This multi-scale decoding strategy further enhances segmentation performance in the \acrshort{bev} domain by leveraging feature representations at different feature resolutions, a principle well established in both medical image segmentation and occupancy grid mapping literature \cite{krithika2022review, li2023fb}. By integrating high-resolution spatial details with deep semantic context, our lightweight U-Net variant achieves precise boundary delineation and robust class discrimination across scales. We selected all feature-map dimensions empirically to ensure that the network remains efficient enough for real-time deployment without sacrificing metric accuracy. Overall, this design balances computational cost, spatial fidelity, and semantic richness, yielding a versatile map-segmentation head that performs reliably under the strict latency constraints of a real-time application.

\subsection{Loss Functions}

To achieve accurate and robust semantic segmentation, we employ a loss set that combines, by a weighted sum, six complementary terms inspired by Li \textit{et al.} \cite{li2023fb}. The loss weights were chosen empirically so that the initial error values extracted from each loss were approximately equal. Rather than relying solely on a single criterion, as we can find in the literature for the map segmentation task, this multi–loss design addresses class imbalance, mask overlap, region consistency, and geometric plausibility. Additionally, we are adding a new loss to this set to address the boundary precision and improve our metrics. The full set of losses will be detailed as follows.

The first loss and the most used in the literature for the map segmentation problem is Focal Loss \cite{lin2017focal}, extending binary cross‐entropy by down‐weighting easy examples and focusing on hard samples. This loss has a focus on individual cell-level optimisation and can be expressed using \eqref{eq:focal_loss}, as follows:
\begin{equation}
\label{eq:focal_loss}
\mathcal{L}_{\text{focal}} = - \frac{1}{C} \sum_{c=1}^{C} \alpha_c \, (1 - p_{i,c})^\gamma \, \log(p_{i,c})
\end{equation}

where $C$ is the number of classes, $p_{i,c}$ denotes the predicted probability for class $c$ at cell $i$, and for the hyperparameters, we are using $\gamma=3.0$ and $\alpha_c=0.25$ to control the focusing and balance, according to \cite{schramm2024bevcar}, and we apply a weight of \(\lambda_{\text{focal}} = 7.0\) to this term to ensure that the starting values of all losses were of similar scale.

The second loss used was Dice Loss \cite{sudre2017generalised}, which directly optimises the overlap between the predicted segmentation map and the ground truth by maximising the Dice coefficient, using a simplified global approach. In our work, we are using the V-Net \cite{milletari2016vnetfullyconvolutionalneural} version. This mapping can be expressed by \(2\times \) the overlap between the predicted probabilities and the ground truth, normalised by the sum of their squared volumes. Formally, it is defined as \eqref{eq:dice_loss}:
\begin{equation}
\label{eq:dice_loss}
\mathcal{L}_{\text{dice}} = 1 - \frac{1}{C} \sum_{c=1}^{C} \frac{2 \sum_i p_{i,c} g_{i,c}}{\sum_i p_{i,c}^2 + \sum_i g_{i,c}^2 + \varepsilon}
\end{equation}

where $C$ is the number of classes, \(p_{i,c}\) and \(g_{i,c}\) denote the predicted probability and the binary ground-truth label for class $c$ at cell $i$, respectively, and \(\varepsilon\) is a small constant added for numerical stability. Although the Dice coefficient uses the sum of predicted and ground truth volumes in the denominator, it does not compute the true set-theoretic union (i.e., $|A| + |B| - |A \cap B|$), which distinguishes it from the Jaccard Index~\cite{sudre2017generalised}. Considering this, Dice loss effectively mitigates class imbalance and encourages the model to produce coherent, high-quality segmentation masks, being widely used in medical image analysis and other data-imbalanced tasks \cite{li2019dice}.  
This term is weighted by \(\lambda_{\text{dice}} = 10.75\) so that the initial magnitudes of each loss were roughly the same.

Our third component is the Lovász-Softmax Loss \cite{berman2018lovasz}, which differs from Dice loss in its optimisation objective. While Dice loss focuses on the overlap between predicted and ground truth masks, Lovász-Softmax directly optimises the \acrfull{iou} metric by leveraging the convex Lovász extension of the Jaccard set function, using a more refined global approach. For each class \(c\), a vector of cell errors \(m^{(c)}\) of length \(N\) (number of discretized cells) is constructed, where
\begin{equation}
m_i^{(c)} =
\begin{cases}
1 - p_{i,c}, & \text{if } g_i = c,\\
p_{i,c},     & \text{otherwise}
\end{cases}
\end{equation}
and \(p_{i,c}\) is the predicted probability for class \(c\) at cell \(i\), and \(g_i\) is the respective ground truth. Sorting \(m^{(c)}\) in descending order yields \(m_{(1)}^{(c)} \ge \cdots \ge m_{(N)}^{(c)}\). The Lovász-Softmax loss is then defined as
\begin{equation}
\label{eq:lovasz_loss}
\mathcal{L}_{\text{lovasz}}
= \frac{1}{C}\sum_{c=1}^C \sum_{i=1}^N m_{(i)}^{(c)} \,\Delta J_i^{(c)}
\end{equation}
where \(\Delta J_i^{(c)}\) is the discrete gradient of the Jaccard index (also known as \acrshort{iou}) \cite{powers2020evaluation} when adding the \(i\)-th largest error back into the prediction set, where \(N\) is the total number of cells, and \(C\) is the total number of classes. This loss provides a continuous, convex approximation whose gradients are aligned with the IoU objective, leading to superior segmentation performance compared to traditional cell-wise losses.  For this term, we are using a weight of \(\lambda_{\text{lovasz}} = 7.5\) so that the initial error values extracted from each loss were approximately equal.

Semantic Scene-Class Affinity Loss is our fourth loss. To explicitly encourage global semantic consistency, we apply the Scene-Class Affinity Loss, defined in MonoScene \cite{cao2022monoscene}, to the semantic predictions \(\hat{y}\) and ground truth labels \(y\). First, for each class \(c\) and cell \(i\), are defined
\begin{align}
\label{eq:affinity_components_1}
P_{c}(\hat p, p) &= \log\frac{\sum_{i} \hat p_{i,c}\,\llbracket p_i = c\rrbracket}{\sum_{i}\hat p_{i,c}}, \\
\label{eq:affinity_components_2}
R_{c}(\hat p, p) &= \log\frac{\sum_{i} \hat p_{i,c}\,\llbracket p_i = c\rrbracket}{\sum_{i}\llbracket p_i = c\rrbracket}, \\
\label{eq:affinity_components_3}
S_{c}(\hat p, p) &= \log\frac{\sum_{i}(1 - \hat p_{i,c})\,(1 - \llbracket p_i = c\rrbracket)}{\sum_{i}(1 - \llbracket p_i = c\rrbracket)}
\end{align}
where \(\hat p_{i,c}\) is the predicted probability that voxel \(i\) belongs to class \(c\) (\(C\) representing the total number of classes), \(p_i\) is its true class, and \(\llbracket\cdot\rrbracket\) is the Iverson bracket. The semantic affinity loss is then
\begin{equation}\label{eq:sem_scal_affinity}
\mathcal{L}_{\text{sem\_scal}}
= -\frac{1}{C}\sum_{c=1}^C \Bigl[P_{c}(\hat y, y) + R_{c}(\hat y, y) + S_{c}(\hat y, y)\Bigr]
\end{equation}

Analogously, to promote global geometric fidelity, we apply the same affinity formulation (this time using occupied and empty cells) to define our fifth loss: Geometric Scene-Class Affinity Loss \cite{cao2022monoscene}. This loss is computed over the predicted and ground-truth geometric occupancy scores, \(\hat{y}_{\mathrm{g}}\) and \(y_{\mathrm{g}}\), respectively. This loss is defined as \eqref{eq:geo_scal_affinity}:
\begin{equation}
\label{eq:geo_scal_affinity}
\mathcal{L}_{\text{geo\_scal}}
= -\frac{1}{C}\sum_{c=1}^C \Bigl[P_{c}(\hat y_{\mathrm{g}}, y_{\mathrm{g}}) + R_{c}(\hat y_{\mathrm{g}}, y_{\mathrm{g}}) + S_{c}(\hat y_{\mathrm{g}}, y_{\mathrm{g}})\Bigr]
\end{equation}
where each term \(P_c\), \(R_c\), and \(S_c\) is computed as in \eqref{eq:affinity_components_1}, \eqref{eq:affinity_components_2}, and \eqref{eq:affinity_components_3} but with geometry labels. By jointly optimizing \(\mathcal{L}_{\text{sem\_scal}}\) and \(\mathcal{L}_{\text{geo\_scal}}\), Scene-Class Affinity Loss improves both semantic correctness and geometric plausibility of the completed scene by supervising local, but contextual consistency between cells belonging to the same object instance or semantic region, enhancing intra-class compactness \cite{cao2022monoscene}. We are using \(\lambda_{\text{sem}} = 1.0\) and \(\lambda_{\text{geo}} = 2.5\) as weights for each one, respectively, so that all losses began with comparable numerical values.

Finally, we are proposing our sixth loss as an adapted version of the Boundary Loss \cite{kervadec2018boundary}, where we are applying a weighted approach to reduce the interference of the background in the distance map. Following Kervadec et al.~\cite{kervadec2018boundary}, we define a distance‐weighted boundary loss that multiplies each predicted probability by its normalised distance to the ground‐truth contour. First, we are applying a normalisation defined by \eqref{eq:bound_normalization}:
\begin{equation}
\label{eq:bound_normalization}
\tilde d_{c,i} =
\begin{cases}
\alpha \cdot d_{i,c},   & \text{if } d_{i,c} > 0, \\
m_{\alpha} \cdot d_{i,c}, & \text{otherwise}
\end{cases}
\end{equation}
where \(\alpha\), and \(m_{\alpha}\) are the scaling parameters for distances outside the object and the corresponding value for the inside region. This formulation allows the model to emphasise either interior or exterior boundary regions, depending on the choice of parameters. For the map segmentation task, we are using only these scaling parameters, with \(\alpha\) and \(m_{\alpha}\) equal to 0.1 and 1.0, respectively. The resulting \(\tilde d_{c,i}\) is then used in the boundary loss definition~\eqref{eq:norm_bound}.
\begin{equation}
\label{eq:norm_bound}
\mathcal{L}_{\text{norm\_bound}}
= \frac{1}{|C|\,|N|}
\sum_{c \in C}
\sum_{i \in N}
p_{i,c}\,\tilde d_{i,c}
\end{equation}
where the predicted class probability \(p_{i,c}\) and the normalized distance \(\tilde d_{i,c}\) to the ground-truth boundary. Here, \(C\) is the total number of classes, and \(N\) is the total number of cells. This emphasises the cells' importance inside the boundaries and reduces the contribution of cells outside the object boundaries, specifically those associated with predictions unrelated to the target classes, encouraging sharper, more accurate object edges in the segmentation map. The final normalised boundary loss is scaled by a weight \(\lambda_{\text{norm\_bound}} = 1.25\) to keep the initial outputs of each loss at a similar magnitude.

The total loss for the map segmentation head is the weighted sum in \eqref{eq:total}, where each loss component is scaled by a dedicated weight that was empirically selected to balance the influence of regional, boundary, and distribution-aware terms, ensuring a stable and effective multi-objective optimisation.
\begin{equation}\label{eq:total}
\begin{split}
\mathcal{L} =\;& 
\lambda_{\text{focal}}\,\mathcal{L}_{\text{focal}}
+ \lambda_{\text{dice}}\,\mathcal{L}_{\text{dice}} + \\&
\lambda_{\text{lovasz}}\,\mathcal{L}_{\text{lovasz}} + \lambda_{\text{sem}}\,\mathcal{L}_{\text{sem\_scal}} + \\&
\lambda_{\text{geo}}\,\mathcal{L}_{\text{geo\_scal}}
+ \lambda_{\text{norm\_bound}}\,\mathcal{L}_{\text{norm\_bound}}
\end{split}
\end{equation}

We defined \(\lambda_{\text{focal}} = 7.0\) to emphasise Focal Loss, which addresses class imbalance and focuses learning on difficult examples. The Dice Loss receives the highest weight, \(\lambda_{\text{dice}} = 10.75\), due to its effectiveness in maximising region-wise overlap. The Lovász-Softmax Loss is weighted by \(\lambda_{\text{lovasz}} = 7.5\) to enhance optimisation concerning the intersection-over-union (IoU) metric. The Semantic Scaling Loss is assigned a low weight of \(\lambda_{\text{sem}} = 1.0\) to encourage minor alignments between predicted and true label distributions, while the Geometric Scaling Loss is weighted with \(\lambda_{\text{geo}} = 2.5\) to promote spatial alignment. Finally, the Boundary Loss is weighted with \(\lambda_{\text{norm\_bound}} = 1.25\), refining contour precision without overpowering other components.

\section{METHODOLOGY}\label{sec:methodology}
In this section, we provide a comprehensive overview of the methodology adopted in this study, including information about the dataset and the main evaluation metrics adopted to benchmark our architecture against existing methods in the literature, to ensure reproducibility and consistency across experiments. We then outline the tools, frameworks, and computational resources employed during model development and evaluation, including software libraries and hardware specifications.

\subsection{DATA AND METRICS}

For this study, the nuScenes dataset \cite{nuscenes2019} was utilised as it provides one of the largest and most comprehensive datasets in the autonomous vehicle domain. It provides synchronised data from multiple sensors, including 1 lidar, 6 cameras, and 5 radars, operating at frequencies of 20 Hz, 12 Hz, and 13 Hz, respectively. These sensors are positioned to capture a full 360-degree field of view, enabling robust perception for complex driving scenarios. In this architecture, we are using only the radars and cameras for predictions, and we are following Sze and Kunze~\cite{sze2024real}, to define real-time operation as a minimum of 20 \acrshort{fps}, which is consistent with the highest data rates of the sensors used in the nuScenes dataset.

In this article, we are primarily evaluating our model performance based on \acrshort{miou}, considering true positive (TP), false positive (FP), and false negative (FN) metrics of each class \(c\), where \(C\) is the total number of classes. This metric can be formulated as follows:

\begin{equation}
\label{eq:miou}
\text{mIoU} = \frac{1}{C} \sum_{c \in C} 
\frac{ \text{TP}_c }
     { \text{TP}_c + \text{FP}_c + \text{FN}_c }
\end{equation}

For training, the complete nuScenes training set was utilised, and for our quantitative results comparing with our baselines, we utilised the full validation set. Additionally, the validation subsets were created by partitioning the validation set using its description in the metadata for rainy and nighttime data. Regarding input resolution, this study adopts an image size of \(256 \times 704\) pixels, which corresponds to approximately 13\% of the original resolution provided by nuScenes (\(1600 \times 900\) pixels). This downsampling choice aligns with recent literature~\cite{li2023bevdepth, liu2022bevfusion, huang2021bevdet, kim2023crn, lin2024rcbevdet}, where smaller image sizes are used to significantly reduce computational overhead and accelerate inference time on the order of 9\(\times\) the full image, while maintaining comparable performance to models trained on full-resolution images.

\subsection{IMPLEMENTATION DETAILS}

The PyTorch library, MMDetection3D toolbox \cite{mmdet3d2020}, and Python language were used to train and validate the model. Additionally, our model was trained and evaluated using a NVIDIA A100 GPU for 24 epochs, using the AdamW optimizer in an end-to-end manner, and the hyper-parameters used during training are as follows: a baseline learning rate of \begin{math} 2e-4 \end{math}, reducing the learning rate by a factor of 0.1 in the 5th last and last epochs. In addition, our model was trained using a map grid of \(200\times200 \), with \(0.5\) m of resolution and using the previous six radar sweeps.

Regarding the computational resources needed to evaluate our architecture, we used the Pytorch memory tools to measure the VRAM usage, psutil (process and system utilities) to measure the RAM usage, and 1 as batch size to perform the evaluation and memory analysis. According to our analysis, the maximum VRAM allocated was 17.90 GB, and the RAM usage was 6.63 GB during the execution of our process. Additionally, our model contains 58.4 million parameters, and the inference time reported in our experimental results was measured using 100 warm-up iterations, followed by the average inference time of the subsequent 900 iterations.

\section{RESULTS}\label{sec:results}
In this section, we report the experimental results obtained with our proposed approach, highlighting its performance in comparison to state-of-the-art techniques. We then conduct an ablation study to assess the individual impact of key components in our model and to gain a deeper understanding of their contributions to the overall system performance. Finally, we provide a qualitative analysis, showcasing visual examples that demonstrate how our model enhances map segmentation quality in complex urban scenes, which are representative of real-world autonomous driving scenarios.

\subsection{EXPERIMENTAL RESULTS}

To assess the performance of the proposed architecture in the map segmentation task, we conducted experiments comparing the results with some baseline models, mainly UniM$^2$AE \cite{zou2024unim} and BEVCar \cite{schramm2024bevcar} as references. The analysis primarily focuses on the \acrshort{miou}, considering all available classes, which is the standard evaluation metric for map segmentation in the nuScenes benchmark. Table~\ref{tab:main_results} summarises our main results, with the row corresponding to our model highlighted in grey.
% A key result is that our approach achieved an mIoU of 53.5, outperforming the baseline despite employing a smaller image backbone.

\begin{table*}[!htbp]
    \renewcommand{\arraystretch}{1.2}
    \caption{Comparison of different map segmentation methods on the validation set. Highlighted values in bold represent the best performance in the respective category. Additionally, C and R represent camera and radar, respectively. We are reporting the results found in each respective paper, where the unreported details are represented by $-$.}
    \label{tab:main_results}
    \setlength{\tabcolsep}{5.4pt}
    \begin{tabular}{l l l c c c c c c c}
    \hline
    Method & Modalities & Image Backbone & Drivable $\uparrow$ & Ped. Cross.$\uparrow$ & Sidewalk $\uparrow$ & Stop Line $\uparrow$ & Carpark $\uparrow$ & Divider $\uparrow$ & mIoU $\uparrow$ \\ 
    \hline
    CVT \cite{zhou2022cross} & C & EfficientNet & 74.3 & 36.8 & 39.9 & 25.8 & 35.0 & 29.4 & 40.2 \\
    OFT \cite{roddick2018orthographic} & C & - & 74.0 & 35.3 & 45.9 & 27.5 & 35.9 & 33.9 & 42.1 \\
    LSS \cite{philion2020lift} & C & ResNet-50  & 75.4 & 38.8 & 46.3 & 30.3 & 39.1 & 36.5 & 44.4 \\
    M$^2$BEV \cite{xie2022m} & C & ResNet-50  & 77.2 & – & – & - & – & 40.5 & – \\
    BEVFusion \cite{liu2022bevfusion} & C & Swin-T & 78.2 & 48.0 & 53.5 & 40.4 & 45.3 & 41.7 & 51.2 \\
    UniM$^2$AE \cite{zou2024unim} & C & Swin-T & 79.5 & 50.5 & 54.9 & 42.4 & 47.3 & 42.9 & 52.9 \\
    BEVFormer-S \cite{li2022bevformer} & C & ResNet-101 & 80.7 & – & – & - & – & – & – \\
    Simple-BEV++ \cite{harley2023simple} & C+R & ViT-B/14 & 81.2 & – & – & - & – & – & 50.4 \\
    BEVGuide \cite{man2023bev} & C+R & EfficientNet & 76.7 & – & – & - & – & – & – \\
    BEVCar \cite{schramm2024bevcar} & C+R & ResNet-101 & \textbf{81.8} & – & – & - & – & –& 53.0 \\
    \rowcolor[gray]{0.9}
    FIN (Ours) & C+R & ResNet-50 & 
    67.7 & \textbf{53.1} & \textbf{60.7} & \textbf{45.6} & \textbf{51.3} & \textbf{48.2} & \textbf{53.5} \\
    \hline
    \end{tabular}
\end{table*}

Analysing the results in Table~\ref{tab:main_results}, it can be seen that our method, FIN, achieves the highest overall performance in terms of \acrshort{miou} (53.5), surpassing SOTA approaches, including those using more complex image backbones such as ResNet-101 and ViT-B/14. Additionally, unlike the baseline models, which show much better results for the drivable surface, the most common class, compared to the overall \acrshort{miou}, our method provides detailed class-wise results that demonstrate strong consistency across categories. These results highlight the effectiveness of our architecture not only in overall accuracy but also in capturing fine-grained map elements relevant to autonomous driving scenarios.

\begin{figure}[!htbp]
\centering
\includegraphics[width=3.3in]{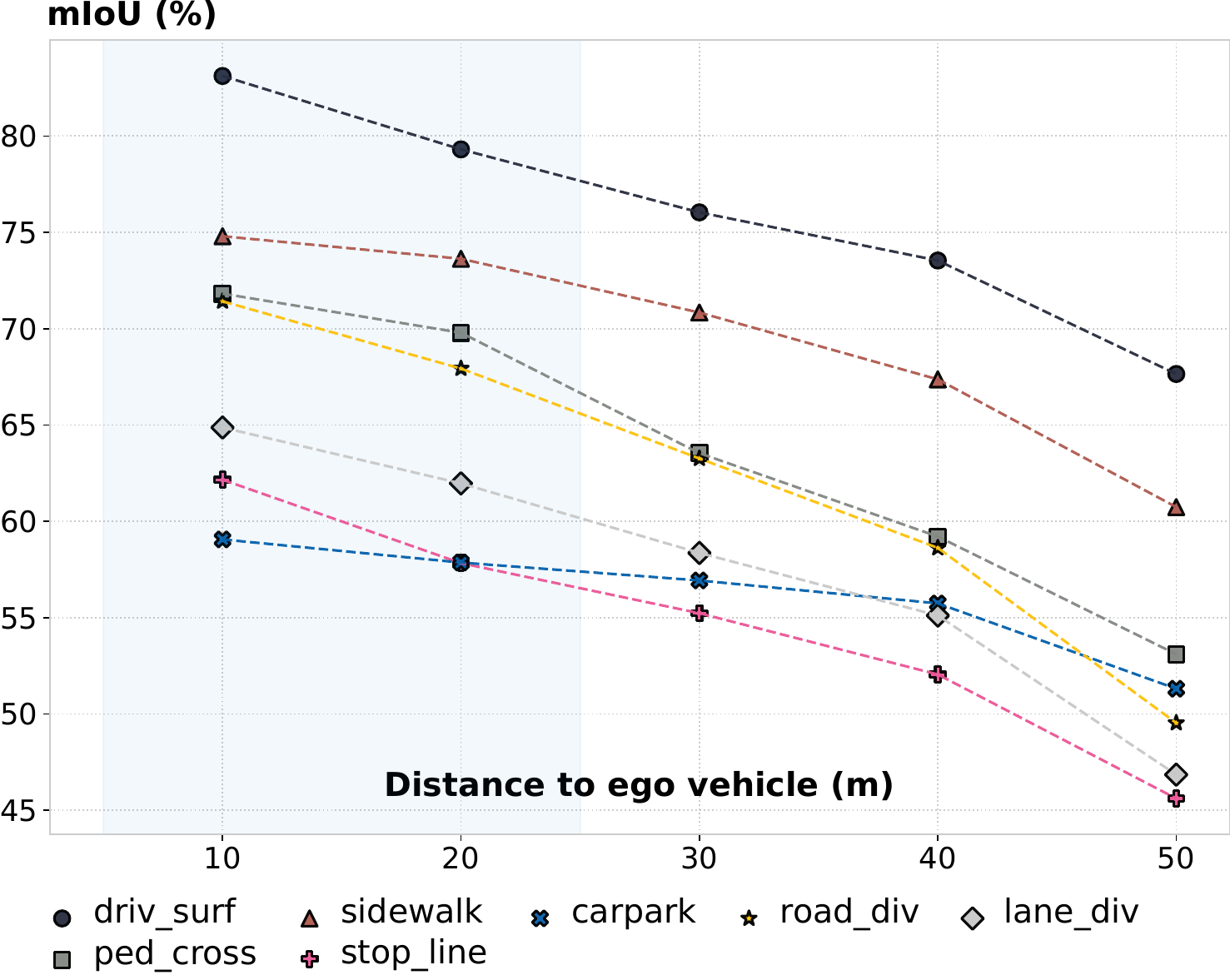}
\caption{\acrshort{miou} degradation per class under different distances to the ego vehicle. In this experiment, the full validation set was used. The blue area [0,25] represents the safe driving distance considering the speed limit of 50 km/h and 1 s of reaction time.}
\label{fig:miou_distance_classes}
\end{figure}

Additionally, we validated our approach across all map segmentation classes by splitting the "divider" category into "lane divider" and "road divider" and by evaluating performance at varying distances from the ego vehicle, ranging from 10 m to 50 m, the standard evaluation distance presented in the literature, in 10 m increments. The primary aim of this experiment is to quantify how \acrshort{miou} degrades with increasing cell distance, and the results are presented in Figure~\ref{fig:miou_distance_classes}.

From the distance to ego vehicle analysis, considering all classes for the map segmentation task, we can observe a proportional degradation for each class. This behaviour is important to show how our model performs at closer distances for safety aspects and its consistency between classes. Based on \cite{bispo2025panpillarsattentionbasednetwork3d}, we added a distance limit for the desired perfect perception at 25 meters (blue region in the Figure~\ref{fig:miou_distance_classes}), which can be considered the minimum distance at which a self-driving car is expected to have a score as close as possible to 100\% to react in time to avoid accidents. In line with these ideas, it is possible to observe a limitation in terms of reliability in our architecture, despite its presenting good results when compared with the literature.

\begin{figure}[!htbp]
\centering
\includegraphics[width=3.3in]{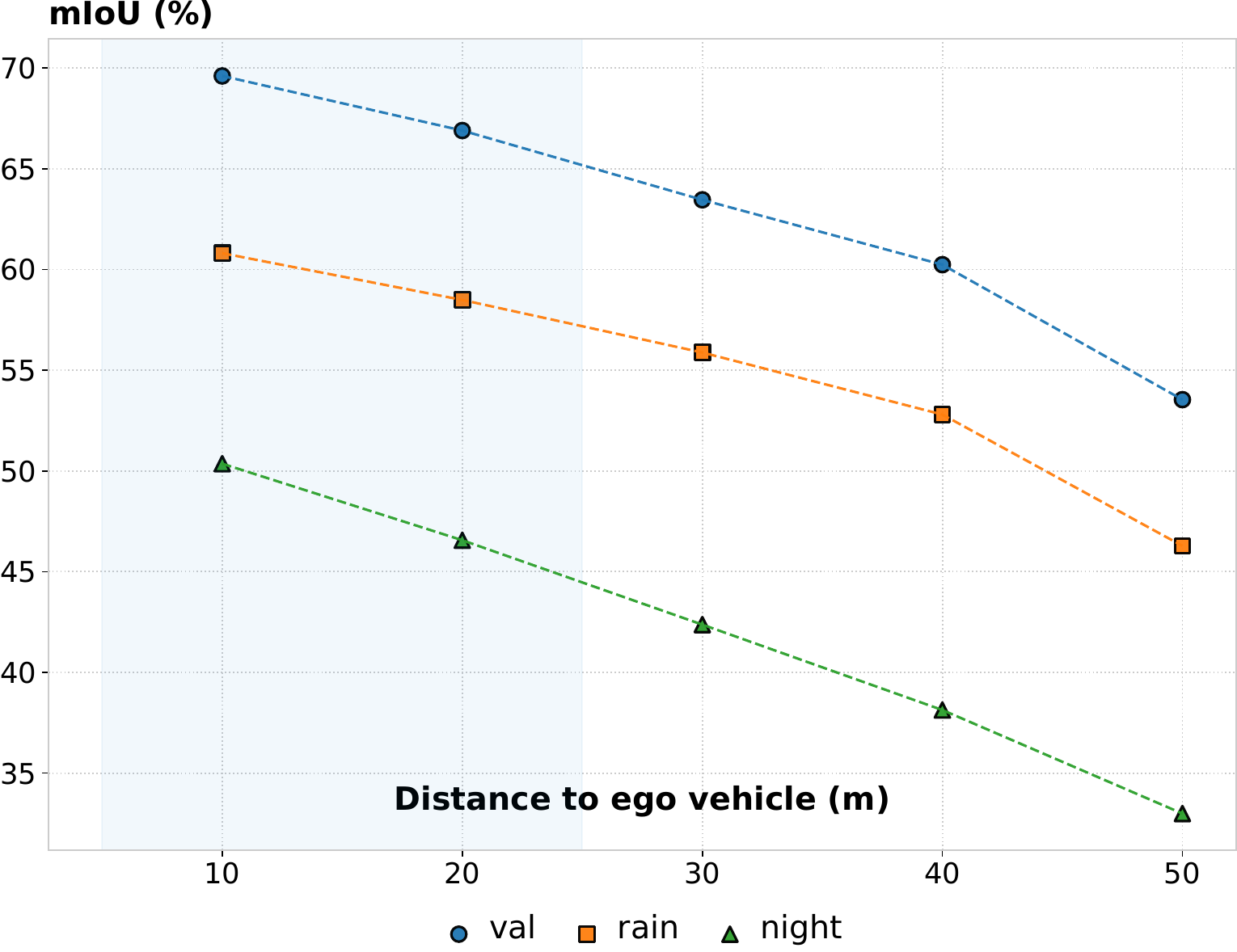}
\caption{\acrshort{miou} degradation per weather condition under different distances to the ego vehicle. The blue area [0,25] represents the safe driving distance considering the speed limit of 50 km/h and 1 s of reaction time.}
\label{fig:miou_distance_valset}
\end{figure}

We further evaluate our model on different validation subsets by splitting the full dataset into two additional subsets, based on the available metadata, to isolate rainy and nighttime scenes. Figure~\ref{fig:miou_distance_valset} presents the analysis of \acrshort{miou} as a function of distance to the ego vehicle for each subset, including the full validation set. As expected, there is a consistent decrease in performance across all three curves, with the worst results observed in the nighttime scenario. This subset contains less semantic information due to the low lighting and poor visibility conditions captured by the camera images, which is crucial to the map segmentation task.

For the runtime analysis, we compared our architecture, \acrshort{fin}, against two well-known models in the literature, Simple-BEV++ and BEVCar, using the same NVIDIA A100 GPU to ensure consistency in evaluation metrics. Both comparison models were selected due to their established results using this specific GPU configuration and their different image backbones, providing a reliable benchmark for performance assessment. The results demonstrate that our model achieves superior efficiency with a significantly lower runtime of 38 milliseconds compared to Simple-BEV++, 277 milliseconds, and BEVCar, 137 milliseconds, delivering higher frames per second at 26.3 FPS versus their respective values of 3.6 FPS and 7.3 FPS, achieving an improvement of about 260\% whereas our model also presents better metrics, in terms of \acrshort{miou}, considering BEVCar as the baseline. These results are presented in Table \ref{tab:runtime_analysis}

\begin{table}[!htbp]
    \centering
    \renewcommand{\arraystretch}{1.2}
    \caption{Runtime analysis of different methods on an NVIDIA A100 GPU. We are showing the BEVCar and Simple-BEV++ results presented by the BEVCar authors.}
    \label{tab:runtime_analysis}
    \setlength{\tabcolsep}{5pt}
    \begin{tabular}{l|c|c|c|c}
        \hline
        Method & Modalities & Image Backbone & 
        Runtime $\downarrow$ & FPS $\uparrow$ \\ 
        \hline
        Simple-BEV++ & C+R & ViT-B/14 & 277 (ms) & 3.6 \\
        BEVCar & C+R  & ResNet-101 & 137 (ms) & 7.3 \\
        \rowcolor[gray]{0.9}
        FIN (Ours)  & C+R & ResNet-50 & 38 (ms) & 26.3 \\
        \hline
    \end{tabular}
\end{table}

\subsection{ABLATION STUDY}

To analyse the effect of different components in our proposed architecture, we conducted key experiments by varying two main components of our model. Firstly, we experiment with the implementation of the head component. We replace our UNet implementation with the widely used convolutional head \cite{schramm2024bevcar}, \cite{zou2024unim}, \cite{liu2022bevfusion}, which is composed of two sets of convolutional layers followed by a normalisation layer and a ReLU activation function, feeding into a final convolutional layer. Secondly, we are comparing our set of losses against the Focal Loss, one of the most common loss functions presented in the literature for this task \cite{lin2017focal}. Additionally, we present a final experiment to illustrate the impact of our proposed modified boundary loss.

In our first experiment, shown in Table~\ref{tab:features}, we present a quantitative comparison between the standard vanilla convolutional head \cite{liu2022bevfusion, schramm2024bevcar, zou2024unim} and our proposed simplified U-Net–based head, both attached to a ResNet‑50 backbone and trained for 24 epochs. For the vanilla head, we used the previously described structure using the feature size of 256 in each convolutional layer, requiring 1.2 million parameters and achieving an \acrshort{miou} of 52.3\%. In the runtime column, we observe that the vanilla head has a shorter run time than the proposed UNet, considering only the head components.
% In the runtime column, it is possible to visualise the runtime, 1.56 ms, considering only this component. 

\begin{table}[h]
    \centering
    \renewcommand{\arraystretch}{1.2}
    \caption{Analysis of our head module using the ResNet-50 version and 24 epochs. The selected heads represent a variation using our proposed simplified U-Net and a Vanilla head, composed of a convolutional layer, which is widely used for \acrshort{bev} architectures in the map segmentation task.}
    \label{tab:features}
    \setlength{\tabcolsep}{6.4pt}
    \begin{tabular}{l|c|c|c}
        \hline
        Component & Parameters $\downarrow$ & mIoU $\uparrow$ & Runtime $\downarrow$ \\
        \hline
        Vanilla Head (Conv2D) & 1.2 M & 52.3 & 1.56 (ms) \\
        U-Net-based Head (Ours) & 1.7 M & 53.5 & 2.14 (ms) \\
        \hline
    \end{tabular}
\end{table}

By contrast, our simplified U-Net module increased the latency to 2.14 ms and the parameter count of the vanilla segmentation head to 1.7 million, achieving an \acrshort{miou} of 53.5\%, representing a relative improvement of around 1.2 points in segmentation accuracy. These results demonstrate that our proposed additional encoder–decoder structure provided by the simplified U-Net head effectively captures richer spatial context in the \acrshort{bev} map segmentation task, while incurring only a modest computational cost increase. 

 The second experiment of the ablation study was based on our set of losses, considering three different scenarios: first, using only the Focal Loss, which yielded 49.6~\acrshort{miou}; second, applying our full loss set with a raw version of the boundary loss, achieving 52.4~\acrshort{miou}; and finally, using the same loss set but replacing the raw boundary term with our proposed normalized boundary loss, which yielded the highest segmentation score. We chose these three configurations because they provide a clear view of how each one of our main proposed components, loss set and loss set considering the boundary loss normalisation, contributes to the final outcome, without resorting to an exhaustive comparison of all possible loss permutations. For this initial assumption, we used a previous work that combined a similar set of losses applied to a different task \cite{li2023fb}. Moreover, this setup highlights that, as shown in Table~\ref{tab:losses}, normalising the boundary term not only promotes more precise edge delineation, which is a critical factor for thin classes, but also boosts average overlap metrics. 

\begin{table}[h]
    \centering
    \renewcommand{\arraystretch}{1.1}
    \caption{Comparison using the ResNet-50 version and 24 epochs. The selected losses represent a variation using a well-established loss for the map segmentation task, our proposed loss set considering the raw boundary loss, and our additional loss normalisation.}
    \label{tab:losses}
    \setlength{\tabcolsep}{13.6pt}
    \begin{tabular}{l|c}
        \hline
        Loss & mIoU $\uparrow$ \\
        \hline
        \begin{math}
            \mathcal{L}_{focal}
        \end{math} & 46.3 \\
        \begin{math}
            \mathcal{L}_{focal+dice}
        \end{math} & 47.8 \\
        \begin{math}
            \mathcal{L}_{focal+dice+lovaz}
        \end{math} & 49.8 \\
        \begin{math}
            \mathcal{L}_{focal+dice+lovaz+(sem\_scal+geo\_scal)}
        \end{math} & 50.0 \\
        \hline
        \begin{math}
            \mathcal{L}_{set}
        \end{math} (all 6 losses) & 52.4 \\
        \begin{math}
            \mathcal{L}_{set}
        \end{math} (all 6 losses) \(+\) boundary loss normalization & 53.5 \\
        \hline
    \end{tabular}
\end{table}

From Table~\ref{tab:losses}, $\mathcal{L}_{set}$ comprises all six loss terms introduced in Section~\ref{sec:architecture}: Focal loss ($\mathcal{L}_{focal}$), Dice loss ($\mathcal{L}_{dice}$), Lovász‐Softmax loss ($\mathcal{L}_{lovasz}$), Affinity loss ($\mathcal{L}_{sem\_scal}$ and $\mathcal{L}_{geo\_scal}$), and Boundary loss ($\mathcal{L}_{boundary}$). Additionally, the final row of Table~\ref{tab:losses} shows the same loss set as the previous row, but with the raw boundary loss replaced by our proposed normalised version via the normalisation function in \eqref{eq:bound_normalization}, as defined in \eqref{eq:norm_bound}.

The results show that using only one loss, as presented in most of the architectures in the literature, composed of either focal loss or cross-entropy loss, the \acrshort{miou} is worse than using our proposed set of losses by 2.8 points. However, if considering our modified boundary loss, applying the distance normalisation to reduce the importance of the background, where the distance to the boundaries is positive, we can improve this score by more than 1.1 points of \acrshort{miou}. This behaviour is consistent with the expected shapes in the map segmentation task, which involves thin structures such as dividers, some roads' boundaries, and sidewalks.

\subsection{QUALITATIVE RESULTS}

In this subsection, we present our qualitative results by comparing \acrshort{fin} map segmentation predictions with the map segmentation ground truth to provide a clear visualization that represents how our model performs under two different scenarios: the first one using the scene 0270 of the validation set, addressing the frame 039, which contains thin classes with irregular shapes; and the second one, using the scene 0625 and frame 039 of the same validation set. This last scene was chosen because it contains a greater variety of shapes for the classes, rainy weather, and low visibility due to occlusions in some regions. 

\begin{figure*}[!ht]
  \centering

  % (a) CRN camera views
  \begin{minipage}[b]{0.8\textwidth}
    \centering
    \includegraphics[width=\linewidth]{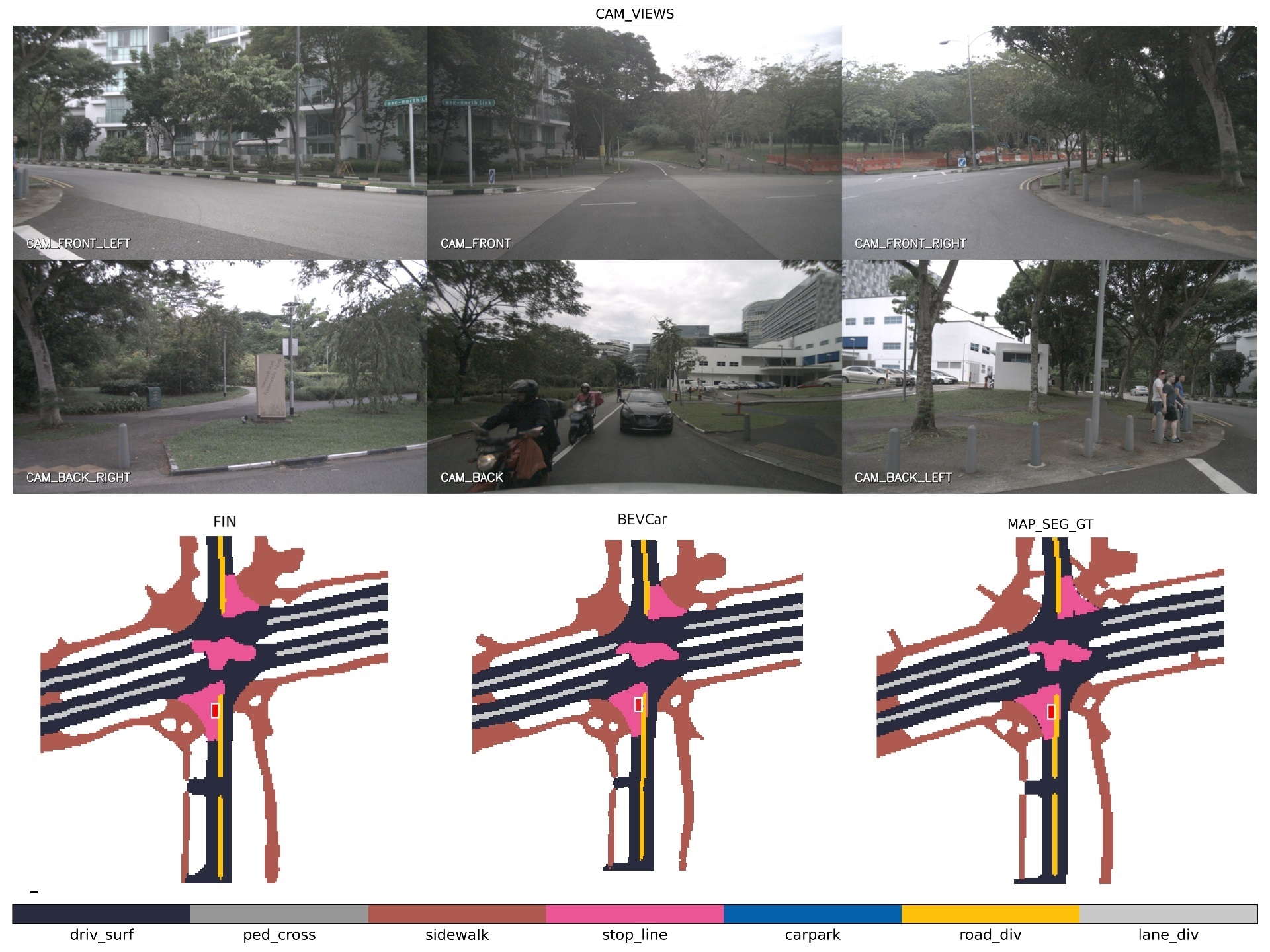}
    \\[-0.3em]
    {\small (a) Scene-0270: Frame-039}
  \end{minipage}

  \vspace{1em}

  % (b) PAN camera views
  \begin{minipage}[b]{0.8\textwidth}
    \centering
    \includegraphics[width=\linewidth]{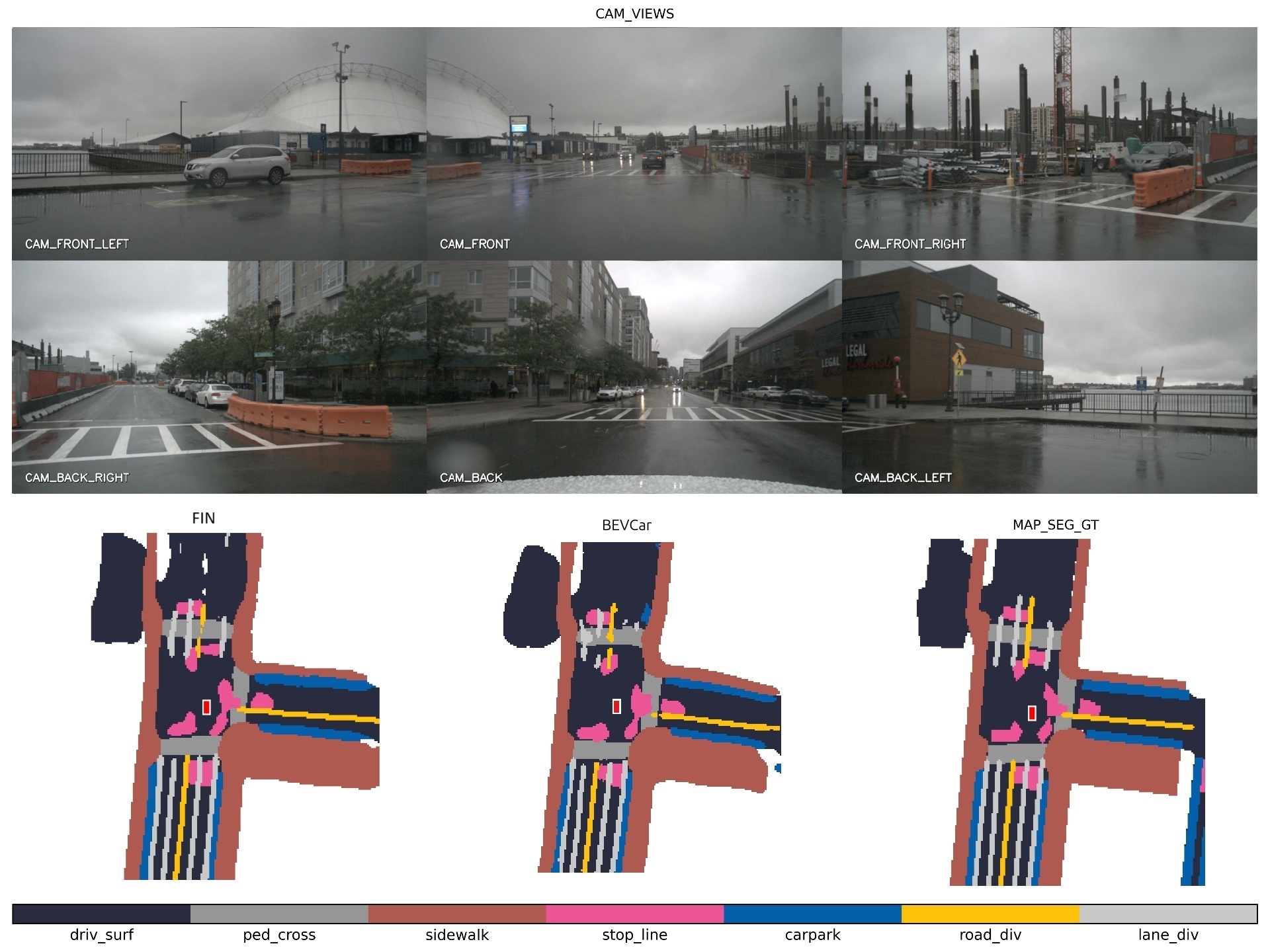}
    \\[-0.3em]
    {\small (a) Scene-0625: Frame-039}
  \end{minipage}

  \vspace{1em}

  \caption{Visualisation for the qualitative results. The results shown in (a) and (b) are our results, FIN, under two different scenes.}
  \label{fig:qual}
\end{figure*}

Fig. \ref{fig:qual} shows those two examples, presenting the 6-camera views, the \acrshort{bev} map segmentation prediction, and \acrshort{bev} map segmentation ground truth. Additionally, the eight possible segmented classes (drivable surface, pedestrian crossing, sidewalk, stop line, carpark, road divider, and lane divider) were coloured according to the presented legend to make this visualisation easier. Finally, each segmented map is centralized in the ego vehicle, represented by a small red rectangle in the center of each \acrshort{bev} map, and the \acrshort{bev} was plotted using the standard distance of -50 m to 50 m for the \(x\) and \(y\) directions in the image, with the resolution of 0.5 m, totalising a 200 m \(\times\) 200 m grid around the ego vehicle.

In Fig. \ref{fig:qual} (a), corresponding to Scene 0270 at Frame 039, our \acrshort{bev} segmentation architecture demonstrates the capability of recovering fine-grained structures even under strong lateral occlusion by vegetation. By employing our composite set of losses, encompassing contour, shape, and positional consistency, we achieve precise delineation of narrow elements such as sidewalks, lane dividers, and road dividers, as well as clear separation of the drivable surface from the background (white areas). Notably, despite the considerable distance from the vehicle and the partial visibility of the frontal field of view, the model maintains coherent and detailed segmentation in regions that would be challenging to infer even for humans.

Additionally, Fig. \ref{fig:qual} (b), for Scene 0625 at the same frame, highlights the robustness of our approach against occlusion by buildings, rainy scenarios, and a complex urban geometry in the background. Even in far-away regions where perspective information is reduced, our model, \acrshort{fin}, with its emphasis on spatial consistency and contour regularisation, faithfully reconstructs road lanes, sidewalks, and intersection layouts. This qualitative performance under adverse visibility conditions underscores the capacity of our model to generalise to distant areas, preserving sharp boundaries and accurately inferring objects at long range, or even highly occluded.

\section{CONCLUSION}\label{sec:conclusion}

In this paper, we propose \acrshort{fin}, a Fast Inference Network for \acrshort{bev} map segmentation, based on camera–radar fusion and evaluated on the nuScenes dataset. We compared our model, which uses ResNet-50 as the image backbone, against models employing stronger backbones such as ResNet-101 and transformer-based architectures. Our approach achieved the best metrics while delivering a substantial gain in inference speed, reaching real-time performance with a 260\% improvement over our most accurate baseline.

To achieve a good balance between metric performance and inference time, we employed a multi-scale head for the map segmentation task, using a simplified U-Net-based architecture to capture features at different scales while maintaining low latency. Moreover, after observing that our initial results could be improved in terms of object shape, we refined the loss function. Our set of losses was modified to incorporate additional contextual and boundary information, enhancing segmentation quality without increasing inference time, since these losses are only applied during training.

Additionally, some limitations of our model remain in the segmentation of occluded and distant objects, which can make it difficult to achieve a perfect score using the standard 200 m \(\times\) 200 m \acrshort{bev} grid. Complementing this analysis, we presented results comparing the \acrshort{miou} as a function of distance from the ego vehicle, which mitigates the impact of occluded and far regions. Our findings suggest that, even for close distances, substantial improvements are necessary to ensure safe driving, which also creates opportunities for subsequent investigations in this field.

Overall, our proposed framework lays a foundation for bridging the gap between map segmentation research and real-world deployment. By achieving real-time inference speeds and high metrics while maintaining a balanced distribution of per-class results, our \acrshort{fin} architecture shows its feasibility for dynamic autonomous driving scenarios. Furthermore, as demonstrated throughout our experiments, the model effectively captures fine-grained spatial details, enabling more precise scene understanding in complex environments. These capabilities position our approach as a practical and scalable solution for real-time autonomous perception systems.

% \section*{APPENDIX}\label{sec:acronyms}
% \printacronyms[style=long]

\bibliography{bib/ref}
\begin{IEEEbiography}[{\includegraphics[width=1in,height=1.3in,clip,keepaspectratio]{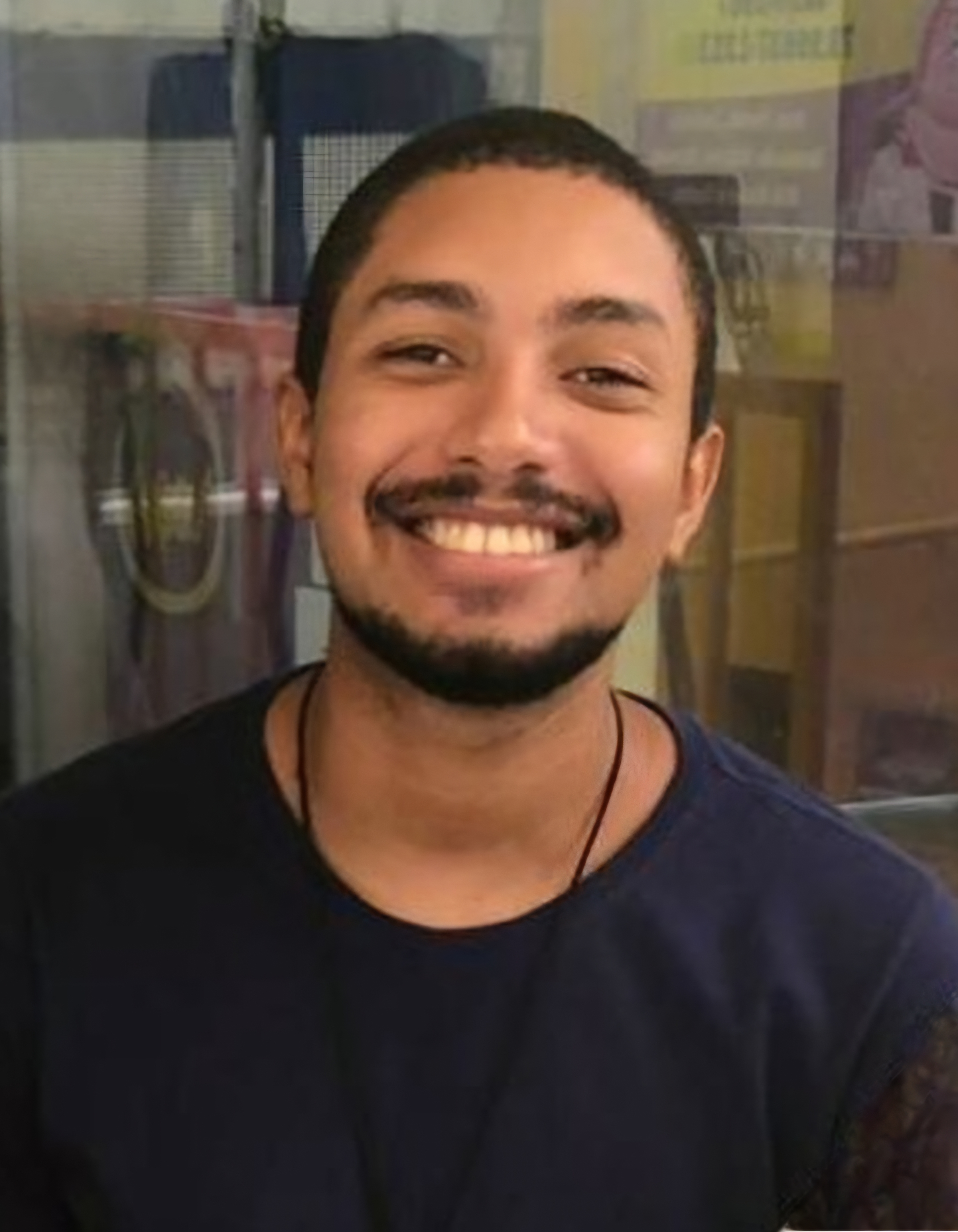}}]{Ruan Bispo } 
received the BA degree in electronic engineering from the Federal University of Sergipe in 2020, the MSE degree in dynamic systems from the University of S\~ao Paulo in 2023, and is currently a second-year PhD student in computer vision and artificial intelligence at the University of Limerick. He was a researcher at Ford Motor Company (2 years) and a data scientist at Samsung (2 years). His research interests include robotics, autonomous vehicles, control, and AI.
\end{IEEEbiography}

\begin{IEEEbiography}[{\includegraphics[width=1in,height=1.3in,clip,keepaspectratio]{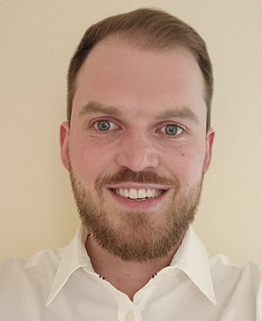}}]{Tim Brophy } 
received the B.Eng. (Hons.) degree from the University of Galway in 2018. From 2018 to 2025, he was a member of the Connaught Automotive Research (CAR) Group at the University of Galway, where he completed his PhD studies. In 2025, he joined the University of Limerick as a Postdoctoral Researcher. His research interests include sensor availability, computer vision, and artificial intelligence in the context of automated vehicles.
\end{IEEEbiography}

\begin{IEEEbiography}[{\includegraphics[width=1in,height=1.3in,clip,keepaspectratio]{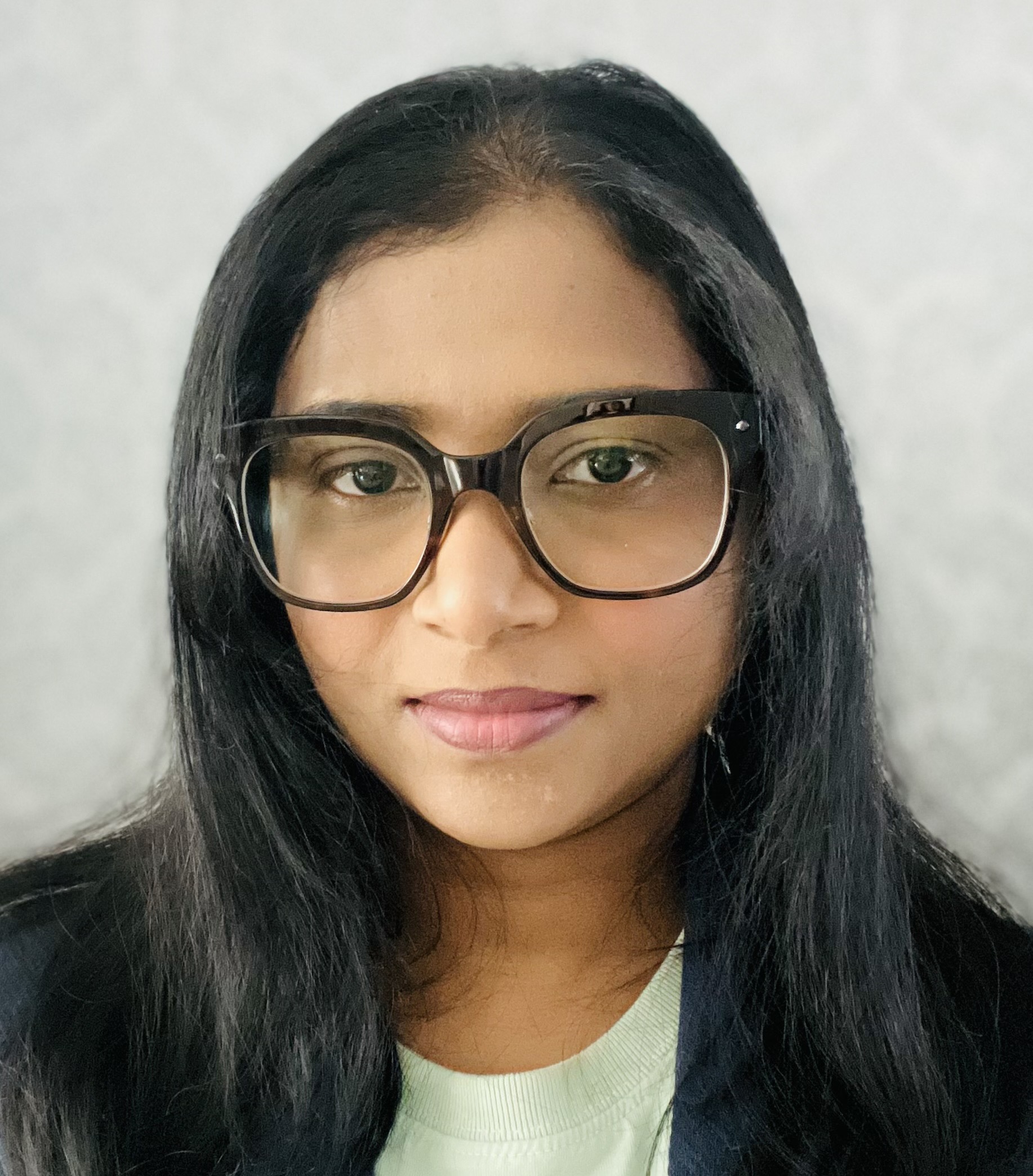}}]{Reenu Mohandas } 
received the M.Tech. degree in digital image computing from the University of Kerala, India, in 2014, the M.Sc. degree in Artificial Intelligence from CIT, currently Munster Technological University, Cork, Ireland, in 2019, and Ph.D. from the Department of Electronic and Computer Engineering (ECE), University of Limerick, Ireland, and currently a Postdoctoral Researcher there. Her research interests include computer vision, deep learning, incremental learning, VQA models, and quantisation and edge deployment of deep learning algorithms.
\end{IEEEbiography}

\begin{IEEEbiography}[{\includegraphics[width=1in,height=1.3in,clip,keepaspectratio]{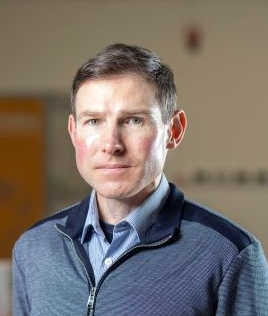}}]{\textbf{Anthony Scanlan }} received the B.Sc. degree in experimental physics from the National University of Ireland Galway, Galway, Ireland, in 1998 and the M.Eng. and Ph.D. degrees in electronic engineering from the University of Limerick, Limerick, Ireland, in 2001 and 2005, respectively. He is currently a Senior Research Fellow at the Dept. of Electronic \& Computer Engineering, University of Limerick, Ireland, and has been the principal investigator on several research projects in the areas of signal processing and data converter design. His current research interests are in the areas of artificial intelligence, computer vision, and industrial and environmental applications. 
\end{IEEEbiography}

\begin{IEEEbiography}[{\includegraphics[width=1in,height=1.3in,clip,keepaspectratio]{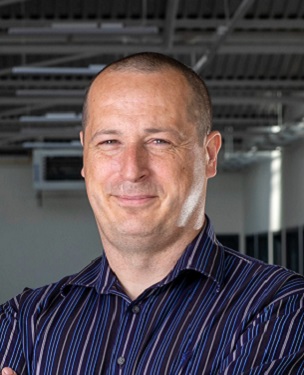}}]{Ciarán Eising } received the BE in Electronic and Computer Engineering and a PhD from the NUI Galway, in 2003 and 2010, respectively. From 2009 to 2020, he was a Computer Vision Architect \& Senior Expert with Valeo. In 2020, he joined the University of Limerick as an Associate Professor.
\end{IEEEbiography}

\end{document}